\documentclass[letterpaper, 10 pt, conference]{ieeeconf}  % Comment this line out if you need a4paper

\IEEEoverridecommandlockouts                              % This command is only needed if 
\usepackage{hyperref}
\hypersetup{
    colorlinks=true,
    linkcolor=blue,
    filecolor=magenta,      
    urlcolor=cyan,
    pdftitle={Your Title},
    pdfpagemode=FullScreen,
}
\usepackage{comment}
\usepackage{booktabs} % For table formatting.
\usepackage{adjustbox} % to resize and center the table
\usepackage{cite}
\usepackage{amsmath,amssymb,amsfonts}
\usepackage[linesnumbered,ruled,vlined]{algorithm2e}
\usepackage{graphicx}
\usepackage{textcomp}
\usepackage{xcolor}
\newcommand{\norm}[2]{\left \lVert #1 \right \rVert_{#2}}
\title{\LARGE \bf
Two-Phase Bilevel Search for the Moving-Target Traveling Salesman Problem with Moving Obstacles
}

\author{Allen George Philip$^{1}$, Anoop Bhat$^{2}$, Sivakumar Rathinam$^{3}$, and Howie Choset$^{2}$% <-this % stops a space
\thanks{$^{1}$Mechanical Engineering, Texas A\&M University,
		College Station, TX 77843-3123.
		Email: {\tt y262u297@tamu.edu}}%
\thanks{$^{2}$Robotics Institute, Carnegie Mellon University, 5000 Forbes                Ave., Pittsburgh, PA 15213, USA.
        Emails: {\tt \{agbhat, choset\}@andrew.cmu.edu}}%
\thanks{$^{3}$Mechanical Engineering, and Computer Science and Engineering,              Texas A\&M University,
		College Station, TX 77843-3123.
        Email: {\tt srathinam@tamu.edu}}
}

\begin{document}

\maketitle
\thispagestyle{empty}
\pagestyle{empty}

\begin{abstract}

The Moving-Target Traveling Salesman Problem (MT-TSP) seeks a minimum cost trajectory for an agent that departs from a static depot, visits a set of moving targets, each within one of their assigned time windows, and returns to the depot. In this article, we study the Moving-Target Traveling Salesman Problem with Moving Obstacles (MT-TSP-MO), a generalization of the MT-TSP where the agent trajectory must avoid moving obstacles. We present a Mixed-Integer Conic Programming (MICP) formulation that can be solved using off-the-shelf solvers, as well as a fast and scalable Two-Phase Bilevel Search (TPBS) algorithm that computes high-quality feasible solutions for the problem. We evaluate our approaches against an existing baseline algorithm on a broad range of problem instances with up to 40 targets and 40 obstacles. The results demonstrate that both the proposed methods significantly outperform the baseline with respect to success rates, solution costs, and computation time.

\end{abstract}

\vspace{2mm}
\section{INTRODUCTION}

Given a set of moving targets with predefined trajectories and associated time windows, the Moving-Target Traveling Salesman Problem (MT-TSP) seeks a minimum-time trajectory for an agent that begins at a specified, fixed location (depot), visits all the targets within one of their respective time windows, and returns to the depot. The MT-TSP has several practical applications including monitoring and surveillance \cite{deMoraes2019, wang2023moving, marlow2007travelling, maskooki2023bi}, resupply missions with mobile vehicles \cite{helvig2003}, missile defense \cite{helvig2003, smith2021assessment, stieber2022}, dynamic target tracking \cite{englot2013efficient}, and industrial robot planning \cite{chalasani1999approximating}. Several algorithms exist for the MT-TSP. Recent works \cite{stieber2022, philip2024mixed, philip2025mixed} provide completeness and optimality guarantees for cases where targets move along linear or piecewise-linear trajectories. The algorithms in  \cite{philip2025c} and \cite{bhat2025parallel} address generic target trajectories by providing tight lower bounds on the optimum and asymptotic guarantees, respectively. Other heuristic approaches include
% \cite{bourjolly2006orbit,choubey2013,deMoraes2019,englot2013efficient,groba2015solving,jiang2005tracking,marlow2007travelling,ucar2019meta,wang2023moving}. 
\cite{bourjolly2006orbit,choubey2013,deMoraes2019,groba2015solving,marlow2007travelling,wang2023moving,englot2013efficient}.

\begin{figure}
    \centering
    \includegraphics[width=0.9\linewidth]{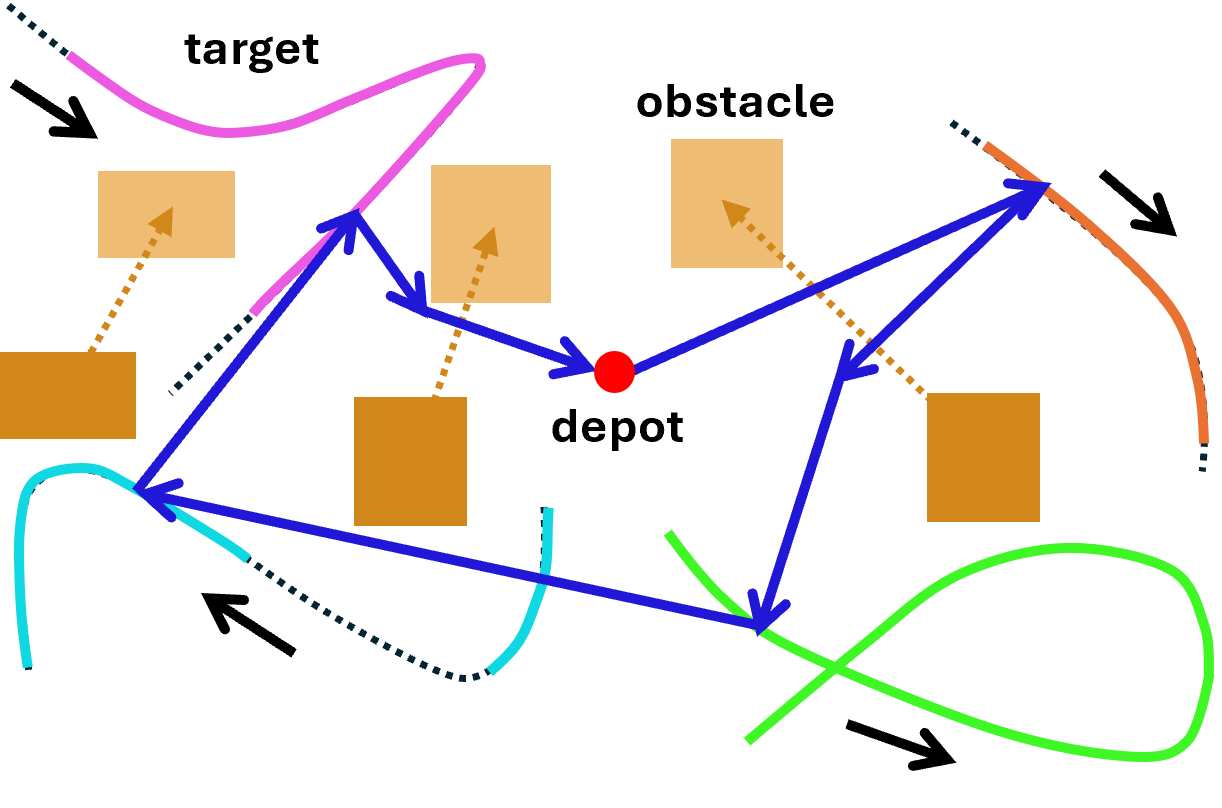}
    \caption{A feasible solution for an example instance of the MT-TSP-MO. Solid, colored portions of the target trajectories indicate their respective time windows. The agent begins and ends its tour at the depot while avoiding moving obstacles.}
    \label{fig:mttspmo}
\end{figure}

The Moving-Target Traveling Salesman Problem with Obstacles (MT-TSP-O) is a natural generalization of the MT-TSP where the agent must avoid static obstacles. Recently, a complete algorithm was proposed for the MT-TSP-O for 2D workspaces \cite{bhat2024complete}, followed by a complete and bounded-suboptimal algorithm for the 3D generalization in \cite{bhat2025complete}. In this article, we consider the Moving-Target Traveling Salesman Problem with Moving Obstacles (MT-TSP-MO) which further generalizes the MT-TSP-O (Fig.~\ref{fig:mttspmo}). Moving-target interception amidst moving obstacles has applications such as underwater replenishment of naval ships \cite{brown2017scheduling} and periodic recharging of Unmanned Aerial Vehicles (UAVs) using Unmanned Surface Vehicles (USVs) \cite{li2023dynamic}. {\color{black} Finding a feasible solution to the MT-TSP-MO is NP-complete \cite{savelsbergh1985local} due to the presence of time windows, making completeness difficult to guarantee.} Furthermore, finding an optimal solution to the MT-TSP-MO is NP-hard as it generalizes the TSP. Currently, the research on the MT-TSP-MO is very limited, with \cite{li2023dynamic} providing the only known approach. Hence, this paper focuses on finding high-quality feasible solutions for the MT-TSP-MO.

{\it Our Contributions:} We first present a Mixed-Integer Conic Programming (MICP) formulation for the MT-TSP-MO. This formulation can be solved using off-the-shelf solvers, and provides asymptotic guarantees on optimality and completeness with respect to a time discretization. However, its scalability is limited for large numbers of targets and obstacles. Hence, we also develop a direct Two-Phase Bilevel Search (TPBS) algorithm that scales to larger problem instances and finds feasible solutions more reliably, at the expense of guarantees on optimality and completeness. TPBS consists of two phases. The first phase quickly generates a set of initial feasible tours, and the second phase improves them to compute lower-cost tours. In both phases, TPBS employs a bilevel search framework that interleaves a high-level search with a low-level search. The {\it high-level search} solves a Generalized Traveling Salesman Problem (GTSP) defined over a target graph consisting of trajectory points for each target sampled within their respective time windows, and the depot. It returns a time-feasible tour for the agent that visits each target within one of its assigned time windows, in the absence of obstacles. The {\it low-level search} then evaluates the feasibility of each edge in the GTSP tour by checking for a collision-free, time-feasible trajectory that connects the corresponding trajectory endpoints. This is accomplished via a fast straight-line check, followed when necessary, by a depth-first search (DFS) on a time-expanded graph that includes obstacle-free workspace locations and target trajectory points at discrete time steps, and the depot.

% \vspace{1mm}
We evaluate the proposed approaches on several problem instances with varying number of targets and obstacles, and total time window durations, and show that they significantly outperform the baseline algorithm in \cite{li2023dynamic} with respect to success rates, solution costs, and computational runtimes. 

% \vspace{-4mm}
\section{PROBLEM DEFINITION}

All the targets, obstacles, and the agent are contained within an $\mathbb{R}^2$ workspace. Let $\mathcal{T} := \{1,2,\cdots,|\mathcal{T}|\}$ denote the set of targets. Each target $i \in \mathcal{T}$ moves along its trajectory $\pi_{i}: [0, T] \rightarrow \mathbb{R}^2$ where $T$ represents the planning horizon. We define the \emph{trajectory point} of target $i$ at time $t$ as 
$\bar{\pi}_i(t) := (i, \pi_i(t), t)$. Associated with target $i$ is also a set of time windows $\mathcal{W}_i := \{[\underline{t}_i^{1}, \bar{t}_i^{1}], [\underline{t}_i^{2}, \bar{t}_i^{2}], \cdots, [\underline{t}_i^{|\mathcal{W}_i|}, \bar{t}_i^{|\mathcal{W}_i|}] \}$ where $\underline{t}_i^{w}$ and $\overline{t}_i^{w}$ denote the start and end time respectively of the $w$th time window for the $i$th target. We consider the depot $s$ to be a static, dummy target with a single time window $[0,T]$. 

Now, let $\mathcal{O} := \{1,2,\cdots,|\mathcal{O}|\}$ denote the set of obstacles. Each obstacle $o \in \mathcal{O}$ is assumed to be a convex polygon whose center moves along a trajectory $\xi_o: [0,T] \rightarrow \mathbb{R}^2$. 
% Let $\mathcal{O}_o \subset \mathbb{R}^2$, defined by the supporting halfspaces $A_op \leq b_o$ denote the region occupied by obstacle $o$ when centered at the origin.
Let $\mathcal{O}_o := \{p \in \mathbb{R}^2 \, | \, A_op \leq b_o\}$ denote the region occupied by obstacle $o$ when centered at the origin.
The time-varying obstacle region is then given by $\mathcal{O}_o(t) := \{p \in \mathbb{R}^2 \, | \, A_o(p - \xi_o(t)) \leq b_o \}$. This work assumes generic non-linear trajectories for the targets and linear trajectories for the obstacles. %Linear obstacle trajectories simplify collision avoidance for the agent. 

Let $\tau_a:[0, t_f] \rightarrow \mathbb{R}^2$ for some $t_f \leq T$ be the agent trajectory. We say $\tau_a$ is feasible if it is speed admissible, i.e. never exceeds the maximum agent speed $v_{max}$ and is collision-free, i.e. never enters the interior of any obstacle. We say $\tau_a$ \emph{visits} a trajectory point $\bar{\pi}_i(t)$ if $\tau_a(t) = \pi_i(t)$, and that $\tau_a$ \emph{follows} a sequence of trajectory points if it visits all the points in the sequence, in order. We define a \emph{tour} as a sequence of trajectory points which takes the form $(\bar{\pi}_s(0), \bar{\pi}_{i_1}(t^{i_1}), \bar{\pi}_{i_2}(t^{i_2}), \cdots, \bar{\pi}_{i_{|\mathcal{T}|}}(t^{i_{|\mathcal{T}|}}), \bar{\pi}_s(t_f))$ where $(i_1, i_2, \cdots, i_{|\mathcal{T}|})$ denotes a permutation of $\mathcal{T}$. A tour is feasible if there exists a feasible agent trajectory that follows the tour, and the associated time $t^{i_\ell}$ for each target $i_\ell$ in the tour lies within one of its associated time windows in $\mathcal{W}_{i_\ell}$. The goal of the MT-TSP-MO is to find a feasible tour $\mathcal{Q}$, a feasible agent trajectory $\tau_a$, and a final time $t_f$ such that $\tau_a$ follows $\mathcal{Q}$ and $t_f$ is minimized.

\section{\MakeUppercase{Mixed-Integer Conic Program (MICP) for MT-TSP-MO}}

We restrict the time window of the depot $s$ to $[0,0]$, and define a depot copy $s'$ with time window $[0,T]$, and require the agent trajectory to start at $s$ and end at $s'$. For convenience, denote the set of nodes $\mathcal{T} \cup \{s,s'\}$ using $\mathcal{T}'$. Let $\mathcal{K} := \{1,2,\dots,|\mathcal{K}|\}$ denote an index set for a set of times $t_k$ that discretize the time horizon $[0,T]$. For each node $i \in \mathcal{T}'$, we denote by $\mathcal{S}_i$ the subset of steps in $\mathcal{K}$ whose associated times $t_k$ lie within one of the time windows associated with $i$. Additionally, for each $k \in \mathcal{K}$, we denote by $\mathcal{V}_k$ the subset of nodes in $\mathcal{T}'$ for which at least one time window contains $t_k$.

% \vspace{1mm}
We now define the decision variables. For each $k \in \mathcal{K}$, the agent position is defined by the variable $p_k \in \mathbb{R}^2$. We require $p_k$ to be bounded by the workspace limits $[p_{min}, p_{max}]$. For each $i \in \mathcal{T}'$ and $k \in \mathcal{S}_i$, we define binary variables $y_{i,k} \in \{0,1\}$ indicating whether the agent visits node $i$ at step $k$. We also introduce binary variables $g_k \in \{0,1\}$ for each $k \in \mathcal{K} \setminus \{1\}$ indicating whether the agent visited the depot copy $s'$ at some step $j < k$. At any step $k \in \mathcal{K}$, the region occupied by obstacle $o \in \mathcal{O}$ is the intersection of a set of halfspaces, defined by the rows $a^{T}_{o,m}$ of $A_o$ and corresponding entries $b_{o,m}$ of $b_o$, where $m \in \mathcal{M}_o := \{1,2,\cdots,|\mathcal{M}_o|\}$ and $|\mathcal{M}_o|$ is equal to the number of rows in $A_o$. The binary variables $h_{o,m,k} \in \{0,1\}$ indicate whether the agent must lie outside the $m$th halfspace. Finally, we define for each $k \in \mathcal{K} \setminus \{|\mathcal{K}|\}$, variable $l_k \geq 0$  denoting the Euclidean distance traveled by the agent between steps $k$ and $k+1$, and auxiliary variable $l^{xy}_k \in \mathbb{R}^2$ representing the agent's displacement from step $k$ to step $k+1$ to help define $l_k$. Our MICP formulation for the MT-TSP-MO is as follows:

%In this section, we present a new Mixed-Integer Conic Programming (MICP) formulation for the MT-TSP-MO. We first state the formulation, and define all notation and other formulation details immediately afterward.

\begin{align}
    & \min \sum_{k \in \mathcal{S}_{s'}} t_k \, y_{s',k} \label{eq:micpobj}
    \intertext{subject to constraints}
    & \sum_{k \in \mathcal{S}_i} y_{i,k} = 1, \;\; \forall \; i \in \mathcal{T}', \label{eq:visitnodeonce} \\ % \cup \{s, s'\}
    & \sum_{i \in \mathcal{V}_k} y_{i,k} \le 1, \;\; \forall \; k \in \mathcal{K}, \label{eq:nodesperstep} \\
    & \sum_{k \in \mathcal{S}_{s'}} t_k \, y_{s',k} \ge \sum_{k \in \mathcal{S}_i} t_k \, y_{i,k}, \;\; \forall \; i \in \mathcal{T}, \label{eq:visitdepcopylast} \\
    & p_k \ge \sum_{i \in \mathcal{V}_k} \pi_i(t_k) y_{i,k} + p_{min}(1 - \sum_{i \in \mathcal{V}_k} y_{i,k}), \; \forall \; k \in \mathcal{K}, \label{eq:posminbigM} \\[1mm]
    & p_k \le \sum_{i \in \mathcal{V}_k} \pi_i(t_k) y_{i,k} + p_{max}(1 - \sum_{i \in \mathcal{V}_k} y_{i,k}), \; \forall \; k \in \mathcal{K}, \label{eq:posmaxbigM} \\
    & g_k = \sum_{j=1}^{k-1} y_{s',j}, \;\; \forall \; k \in \mathcal{K}\setminus\{1\}, \label{eq:defgk} \\
    & p_k \ge \pi_{s'}(t_1)g_k + p_{min}(1-g_k), \;\; \forall \; k \in \mathcal{K} \setminus \{1\}, \label{eq:pintogoal1} \\[1mm]
    & p_k \le \pi_{s'}(t_1)g_k + p_{max}(1-g_k), \;\; \forall \; k \in \mathcal{K} \setminus \{1\}, \label{eq:pintogoal2} %\\
\end{align}
\begin{align}
    & g_k + \sum_{m \in \mathcal{M}_o} h_{o,m,k} = 1, \;\; \forall \; o \in \mathcal{O}, \;\; \forall \; k \in \mathcal{K}\setminus\{1\} \label{eq:relaxobs}, \\[1mm]
    & \sum_{m \in \mathcal{M}_o} h_{o,m,1} = 1, \;\; \forall \; o \in \mathcal{O}, \label{eq:step1hsum} \\[1mm]
    & a^{T}_{o,m}(p_k - \xi_o(t_k)) \ge b_{o,m} - M(1-h_{o,m,k}), \nonumber \\
    & \quad \; \forall \; o \in \mathcal{O}, \;\; \forall \; m \in \mathcal{M}_o, \;\; \forall \; k \in \mathcal{K}\setminus\{|\mathcal{K}|\}, \label{eq:hyperplanesk} \\[1mm]
    & a^{T}_{o,m}(p_{k+1} - \xi_o(t_{k+1})) \ge b_{o,m} - M(1-h_{o,m,k}), \nonumber \\
    & \quad \; \forall \; o \in \mathcal{O}, \;\; \forall \; m \in \mathcal{M}_o, \;\; \forall \; k \in \mathcal{K}\setminus\{|\mathcal{K}|\}, \label{eq:hyperplanesk+1} \\[1mm]
% \end{align}
% \begin{align}
    & l_k^{xy} = p_{k+1} - p_k, \;\; \forall \; k \in \mathcal{K}\setminus\{|\mathcal{K}|\}, \label{eq:deflxy} \\[1mm]
    & l_k^2 \ge \norm{l_k^{xy}}{2}^2, \;\; \forall \; k \in \mathcal{K}\setminus\{|\mathcal{K}|\}, \label{eq:defnorm} \\[1mm]
    & l_k \le v_{max} (t_{k+1}-t_k), \;\; \forall \; k \in \mathcal{K}\setminus\{|\mathcal{K}|\}. \label{eq:timefeas}
    % & y_{i,k} \in \{0,1\}, \;\; h_{o,m,k} \in \{0,1\}, \;\; p_k \in \mathbb{R}^2, \;\; l_k \ge 0, \;\; g_k \in [0,1].
\end{align}

% \vspace{1mm}

% \vspace{1mm}
The objective \eqref{eq:micpobj} minimizes the time at which the agent visits the depot copy $s'$. This objective relies on constraints \eqref{eq:visitnodeonce} and \eqref{eq:visitdepcopylast} which ensure that each node in $\mathcal{T}'$ is visited exactly once, and that $s'$ is visited only after all other nodes, respectively. In addition, constraints \eqref{eq:nodesperstep} limit the agent to visiting at most one node at any step.

The agent's position is coupled to the node assignments through the big-$M$ constraints \eqref{eq:posminbigM} and \eqref{eq:posmaxbigM}, which enforce coincidence between the agent position and the position of node $i$ at step $k$ whenever $y_{i,k} = 1$, and relax this requirement otherwise. The variables $g_k$ which track whether $s'$ has been visited at a previous step are defined through constraints \eqref{eq:defgk}. Following this, big-$M$ constraints \eqref{eq:pintogoal1} and \eqref{eq:pintogoal2} ensure that the agent remains at the depot position at step $k$ if $g_k = 1$, and relax this requirement otherwise. Although \eqref{eq:pintogoal1} and \eqref{eq:pintogoal2} are optional, they can be beneficial as all other constraints are trivially satisfied when they are active. Note that in \eqref{eq:posminbigM}, \eqref{eq:posmaxbigM}, \eqref{eq:pintogoal1}, and \eqref{eq:pintogoal2}, the big-$M$ values chosen are the workspace bounds $p_{min}$ and $p_{max}$.

Now we explain the obstacle avoidance constraints \eqref{eq:relaxobs}--\eqref{eq:hyperplanesk+1}. Constraints \eqref{eq:relaxobs} and \eqref{eq:step1hsum} together require that for any obstacle $o$ and step $k$, at least one of the $\mathcal{M}_o$ hyperplane variables $h_{o,m,k}$ must be active if $s'$ has not been visited yet (i.e. $g_{k}=0$). However, if $g_{k}=1$, then all $\mathcal{M}_o$ of these variables take value 0 and consequently, the obstacle avoidance requirement is relaxed. At each step $k$, the $h_{o,m,k}$ variables for obstacle $o$ are related to the agent position $p_k$ through big-$M$ constraints \eqref{eq:hyperplanesk} and \eqref{eq:hyperplanesk+1} similar to works \cite{ren2025cp}, \cite{afonso2020task}. When $h_{o,m,k} = 1$, \eqref{eq:hyperplanesk} requires $p_k$ to lie outside the halfspace defined by the $m$th inequality for obstacle $o$. However, when $h_{o,m,k} = 0$, this requirement is relaxed. Constraints \eqref{eq:hyperplanesk+1} require that if the agent must lie outside the halfspace defined by the $m$th inequality for obstacle $o$ at step $k$, then this condition must also be satisfied at step $k+1$; otherwise, the condition is relaxed. The purpose of adding \eqref{eq:hyperplanesk+1} is to ensure collision avoidance between the agent and obstacles over the interval between consecutive steps. This holds since obstacles are assumed to follow linear trajectories; consequently, the linear trajectory segment connecting the agent positions at steps $k$ and $k+1$ cannot intersect any obstacle between these steps.

Next, constraints \eqref{eq:deflxy} and \eqref{eq:defnorm} together ensure that the distance $l_k$ traveled by the agent between steps $k$ and $k+1$ satisfies $l_k \ge \norm{p_{k+1}-p_{k}}{2}$. Finally, the time-feasibility constraints \eqref{eq:timefeas} ensure that this distance does not exceed the maximum distance $v_{max}(t_{k+1}-t_k)$ that the agent can travel between consecutive steps $k$ and $k+1$.

From a solution to the MICP, we construct the agent tour $\mathcal{Q}$ based on the values taken by variables $y_{i,k}$, the agent trajectory $\tau_a$ by interpolating between consecutive $(p_k,t_k)$ values, and the final time $t_f$ from the objective value $\sum_{k \in \mathcal{S}_{s'}} t_k \, y_{s',k}$.

% OPTIONAL
% Add a remark on optimal MICP solution approaching MT-TSP-MO optimum as the time discretization approaches infinity.

\section{TWO-PHASE BILEVEL SEARCH}

\begin{figure}
    \centering
    \includegraphics[width=1\linewidth]{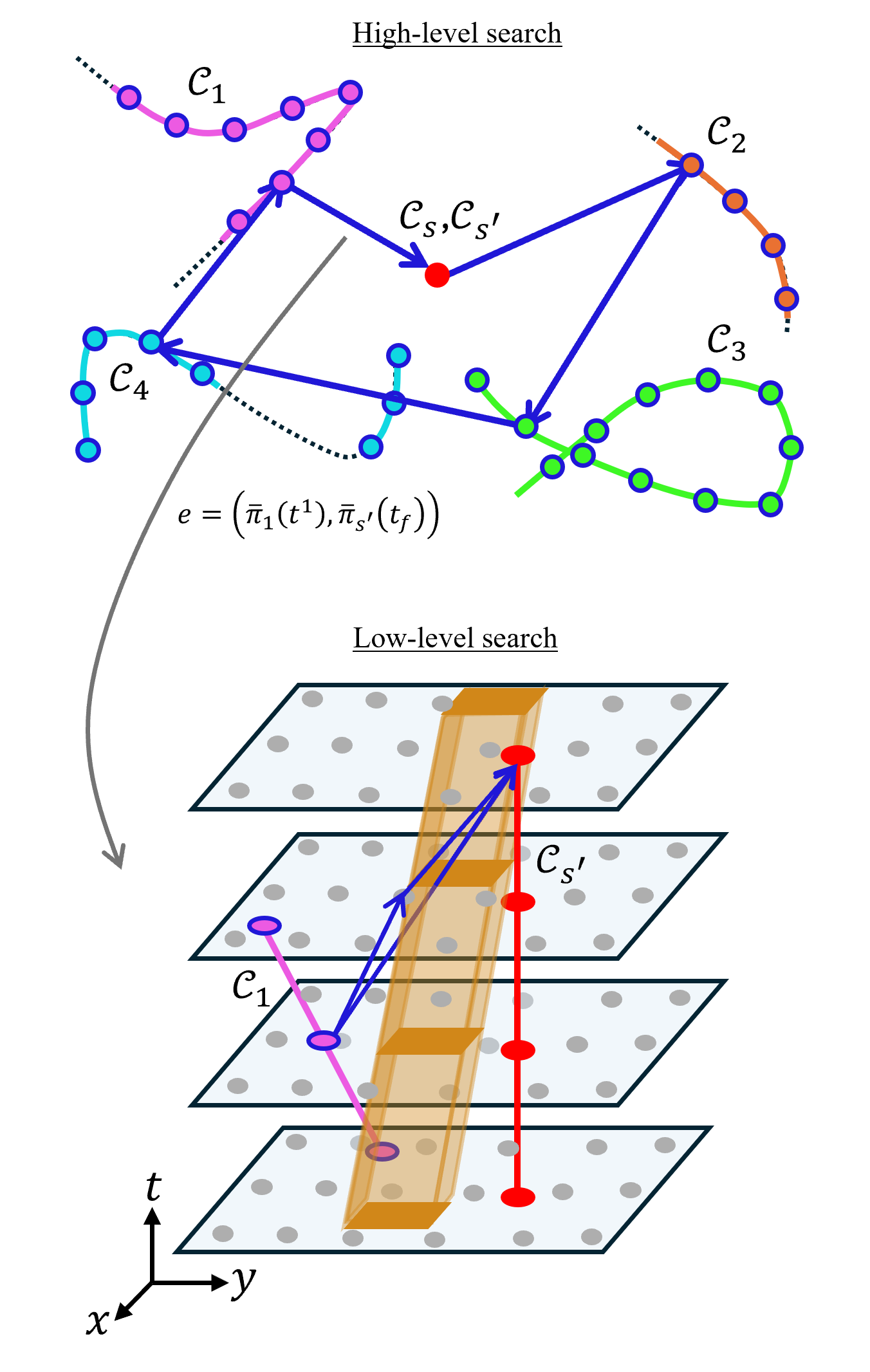}
    \caption{Top: The high-level search returns a time-feasible GTSP tour $\mathcal{Q}$ on $\mathcal{G}_{H}$ that visits the depot and each target once, in the absence of obstacles.
    Bottom: The low-level search evaluates each tour edge by computing a collision-free, time-feasible path in $\mathcal{G}_{L}$ if a straight-line trajectory connecting the edge endpoints is infeasible. In the figure, a straight-line connection for tour edge $(\bar{\pi}_1(t^1), \bar{\pi}_{s'}(t_f))$ collides with an obstacle in between time steps, and is replaced by a feasible path in $\mathcal{G}_{L}$.
    } 
    \label{fig:graphs}
\end{figure}

In this section, we introduce a Two-Phase Bilevel Search (TPBS) algorithm for the MT-TSP-MO. The search consists of a construction phase that quickly generates a set of initial feasible solutions, followed by an improvement phase that refines these solutions. In both phases, TPBS interleaves a high-level and a low-level search. The high-level search solves a Generalized Traveling Salesman Problem (GTSP) on a graph $\mathcal{G}_{H}$ to generate candidate tours, while a low-level search evaluates the feasibility of the tour edges by computing feasible paths in a low-level graph $\mathcal{G}_{L}$ if feasibility cannot be ensured initially through a fast straight-line check. This is illustrated in Fig.~\ref{fig:graphs}, whose notations and further details we will now discuss.

% \vspace{1mm}
Like in the previous section, we set the time window of $s$ to $[0,0]$, introduce a depot copy $s'$ which acts as a static target with associated time window $[0,T]$, and reuse sets $\mathcal{K}, \mathcal{T}', \mathcal{S}_i \textnormal{ for all } i \in \mathcal{T}'$, and $\mathcal{V}_k \textnormal{ for all } k \in \mathcal{K}$. Corresponding to each set $\mathcal{S}_i$, we define a \emph{cluster} $\mathcal{C}_i := \{\bar{\pi}_i(t_k) \, | \, k \in S_i\}$. The set of nodes of $\mathcal{G}_{H}$ is then defined as $\mathcal{V}_{H} := \bigcup_{i \in \mathcal{T}'}\mathcal{C}_i$. The edge set $\mathcal{E}_{H}$ consists of directed edges between all pairs of nodes in $\mathcal{V}_{H}$. Associated with each edge $e$ is a cost ${\tt Cost} (e,\mathcal{G}_{H})$ which takes value $+\infty$ \footnote{From an implementation point of view $+\infty$ is represented using a sufficiently large constant.} if $e$ is infeasible and a finite value otherwise. An edge is marked infeasible if it (a) connects two nodes belonging to the same cluster, or (b) connects any node outside of the depot copy cluster $\mathcal{C}_{s'}$ to the single depot node $\bar{\pi}_{s}(0)$ in $\mathcal{C}_s$, or (c) connects any node from $\mathcal{C}_{s'}$ to nodes outside of $\mathcal{C}_s$. Additionally, an edge $e$ connecting node $\bar{\pi}_i(t_p)$ to node $\bar{\pi}_j(t_q)$ is marked infeasible if $\norm{\pi_j(t_q)-\pi_i(t_p)}{2} > v_{max}(t_q - t_p)$, indicating the agent cannot depart from $i$ at time $t_p$ and arrive at $j$ at time $t_q$ even when traveling at its maximum speed $v_{max}$ while ignoring obstacles. Otherwise, ${\tt Cost} (e,\mathcal{G}_{H})$ is set to $t_q - t_p$. Finally, for all edges $e$ connecting nodes in $\mathcal{C}_{s'}$ to the depot node in $\mathcal{C}_s$, ${\tt Cost} (e, \mathcal{G}_{H})$ is set to 0.

% \vspace{1mm}
We define $\mathcal{G}_{L}$ as a time-augmented graph like in \cite{zhao2025cb}. For each step $k$ in $\mathcal{K} \setminus \{1\}$, we define two sets of nodes, $\mathcal{V}^{free}_k$ and $\mathcal{X}^{free}_k$. We define $\mathcal{V}^{free}_k$ as $\bar{\pi}_i(t_k)$ for all $i \in \mathcal{V}_k$ if $\pi_i(t_k)$ is obstacle-free, and $\mathcal{X}^{free}_k$ as a set of points $(p,t_k)$ in space-time with locations in the obstacle-free regions of the workspace at time $t_k$. The set of nodes of $\mathcal{G}_{L}$ is defined as $\mathcal{V}_{L} := \bar{\pi}_s(0) \, \cup \, \bigcup_{k \in \mathcal{K} \setminus \{1\}} \mathcal{V}^{free}_k \cup \mathcal{X}^{free}_k $. Note that for step $k=1$, we only consider the trajectory point corresponding to the depot at the start time 0 provided that its position is obstacle-free. An edge $e$ connects nodes $v_k$ and $v_{k+1}$ with associated steps $k$ and $k+1$ respectively, if (a) for every obstacle $o \in \mathcal{O}$, there exists at least one inequality $m \in \mathcal{M}_o$ such that the positions associated with both $v_k$ and $v_{k+1}$ lie outside the halfspace defined by $m$ at times $t_k$ and $t_{k+1}$, respectively and (b) the Euclidean distance $d$ between the positions associated with $v_k$ and $v_{k+1}$ satisfies $d \le v_{max}(t_{k+1}-t_k)$. Condition (a) ensures collision avoidance at discrete time steps and along the continuous trajectory between them, and (b) ensures time feasibility. When constructing $\mathcal{G}_{L}$, the edge set $\mathcal{E}_{L}$ is initially empty, with edges generated and added incrementally during the low-level search, according to conditions (a) and (b).

% \vspace{1mm}
% OPTIONAL

% Write a proposition showing how a feasible TPBS solution corresponds to a feasible MICP solution.

% Graph $\mathcal{G}_{H}$ by construction encodes constraints \eqref{eq:visitnodeonce} and \eqref{eq:visitdepcopylast} of the MICP by ensuring that a feasible GTSP tour defined on it begins at the depot cluster, visits all the target clusters exactly once, and finally visits the depot copy cluster.
% Conditions (a) and (b) encode constraints \eqref{eq:hyperplanesk}--\eqref{eq:timefeas} of the MICP directly into the edge feasibility rules of the $\mathcal{G}_{L}$.

% \clearpage

\begin{algorithm}[tb!]
    \label{alg:TPBS}
    \DontPrintSemicolon
    \SetKw{Break}{break}
    \SetKw{Continue}{continue}
    \SetKw{True}{true}
    \caption{Two-Phase Bilevel Search}
    $\mathcal{G}_{L} \gets$ ${\tt ConstrLowLevelGraph}$ () \label{alg:line:GL:TPBS} \;
    $\mathcal{G}_{H} \gets$ ${\tt ConstrHighLevelGraph}$ () \label{alg:line:GH:TPBS} \;
    $\mathcal{F}_{init}, \; \mathcal{F}_{imp} \gets \emptyset, \; \emptyset$ \label{alg:line:makeF:TPBS} \;
    $\mathcal{P}_{H}[e] \gets 0 \;\; \forall \; e \in \mathcal{E}_{H}$ \label{alg:line:makePenalties:TPBS} \;
    \For{\textnormal{trial in} $\{1,2,\cdots,N_{tours}\}$}{\label{alg:line:Phase1start:TPBS}
    \While{\textnormal{initial search time limit not reached}}{
    $\mathcal{Q} \gets$ ${\tt HighLevelDFS}$ $(\mathcal{G}_{H}, \mathcal{P}_{H}, \mathcal{F}_{init})$ \;
    \If {$\mathcal{Q} \textnormal{ is NULL}$}
    {\Break \tcp{end Phase 1} \label{alg:line:endPhase1:TPBS}}
    $\mathcal{E}_{tour} \gets$ ${\tt TourEdges}$ $(\mathcal{Q})$ \;
    \For{\textnormal{each edge $e \in \mathcal{E}_{tour}$}} {
    ${\tt EvalEdge}$ $(e, \mathcal{G}_{L}, \mathcal{G}_{H})$ \;
    }
    \If{\textnormal{all edges in $\mathcal{E}_{tour}$ are feasible after eval}}{
    $\mathcal{F}_{init} \gets \mathcal{F}_{init} \cup \{\mathcal{Q}\}$ \;
    \For{\textnormal{each} $e \in \mathcal{E}_{tour}$} {
    $\mathcal{E}_{win} \gets$ ${\tt WinEdges}$ $(e, \mathcal{E}_{H})$ \;
    $\mathcal{P}_{H}[e_w] \gets \mathcal{P}_{H}[e_w] + \delta \;\; \forall \; e_w \in \mathcal{E}_{win}$}
    \Break \label{alg:line:Phase1end:TPBS}
    }
    }
    }
    \textnormal{Sort $\mathcal{F}_{init}$ in increasing order of tour durations $t_f$} \; \label{alg:line:Phase2start:TPBS}
    \For {\textnormal{tour $\mathcal{Q}$ in $\mathcal{F}_{init}$}}{
    \While{\textnormal{total time limit not reached}}{
    $t_f \gets {\tt time}({\tt last}(\mathcal{Q}))$ \;
    $\mathcal{Q}' \gets$ ${\tt ImproveTour}$ $(\mathcal{Q}, \mathcal{G}_{H})$ \;
    $\mathcal{E}_{tour} \gets$ ${\tt TourEdges}$ $(\mathcal{Q}')$ \;
    \For{\textnormal{each edge $e \in \mathcal{E}_{tour}$}} {
    ${\tt EvalEdge}$ $(e, \mathcal{G}_{L}, \mathcal{G}_{H})$ \label{alg:line:eval2:TPBS} \;
    }
    \If{$\sum_{e \in \mathcal{E}_{tour}} {\tt Cost}(e,\mathcal{G}_{H}) < t_{f}$}{
    $\mathcal{Q} \gets \mathcal{Q}'$
    }
    \If{\textnormal{no new edges evaluated this iteration}}{
    $\mathcal{F}_{imp} \gets \mathcal{F}_{imp} \cup \{\mathcal{Q}\}$ \;
    \Break \label{alg:line:Phase2end:TPBS}
    }
    }
    }
    Find best tour $\mathcal{Q}^{*}$ from $\mathcal{F}_{init} \cup \mathcal{F}_{imp}$ and its cost $t^{*}_f$ \;
    $\tau^{*}_a \gets {\tt GenAgentTraj} (\mathcal{Q}^{*},\mathcal{G}_{L})$ \;
    \Return{\textnormal{$\mathcal{Q}^{*}, \tau^{*}_a, t^{*}_f$}}
\end{algorithm}

\begin{algorithm}[t!]
    \label{alg:highLevelDFS}
    \DontPrintSemicolon
    \SetKw{Continue}{continue}
    \caption{${\tt HighLevelDFS} (\mathcal{G}_{H}, \mathcal{P}_{H}, \mathcal{F}_{init})$}
    BEFORE = dict() \;
    \For{$v \in \mathcal{V}_{H}$}{
    BEFORE[$v$] $\gets$ $\{i \in \mathcal{T}' \, | \, v \notin \mathcal{C}_i \;\textnormal{and } \forall \, v' \in \mathcal{C}_i, \; (v,v') \textnormal{ is marked infeasible} \}$ \;
    }
    STACK $\gets$ [$(\emptyset, \bar{\pi}_s(0))$] \;
    CLOSED $\gets$ $\emptyset$ \;
    \While{\textnormal{STACK is not empty}}{
    $u = (\mathcal{U},v)$ $\gets$ STACK.pop() \;
    \lIf{$u \in \textnormal{CLOSED}$}{\Continue}
    CLOSED.insert($u$) \;
    \If{$\mathcal{U} = \mathcal{T}'$}{\label{alg:line:termstart:highDFS}
    $\mathcal{Q} \gets {\tt ReconstructTour} (u)$ \;
    \lIf{$\mathcal{Q} \textnormal{ does not exist in } \mathcal{F}_{init}$}{\Return{$\mathcal{Q}$}}
    \lElse{\Continue} \label{alg:line:termend:highDFS}
    }
    \For{$v' \textnormal{ in } {\tt SuccessorGNodes} (v, \mathcal{G}_{H}, \mathcal{P}_{H}$)}{\label{alg:line:succ:highDFS}
    $u' = (\mathcal{U} \cup \{{\tt target} (v')\}, v')$ \;
    \lIf{$u' \in \textnormal{CLOSED}$}{\Continue}
    $u'.\textnormal{bp} = u$ \;
    $\textnormal{STACK.push}(u')$ \;
    }
    }
    \Return{\textnormal{NULL}}
\end{algorithm}

% \vspace{1mm}
Now we explain the details of the Two-Phase Bilevel Search (Alg \ref{alg:TPBS}). Lines \ref{alg:line:GL:TPBS}--\ref{alg:line:makeF:TPBS} construct both high and low level graphs, as well as initialize sets that store the initial and improved feasible tours, $\mathcal{F}_{init}$ and $\mathcal{F}_{imp}$, respectively. In line \ref{alg:line:makePenalties:TPBS}, $\mathcal{P}_{H}$ denotes a dictionary mapping each edge in $\mathcal{E}_{H}$ to an associated penalty $\mathcal{P}_{H}[e]$. 
\begin{comment}
$P_{\mathcal{H}}$ denotes a set of penalties associated with edges in $\mathcal{E_H}$, with $P_{\mathcal{H}}(e)$ denoting the penalty for a specific edge $e$.
\end{comment}
All penalties are initially set to 0 in this line. 

% \vspace{1mm}
{\it Phase 1} of the algorithm is described by lines \ref{alg:line:Phase1start:TPBS}--\ref{alg:line:Phase1end:TPBS}. This phase attempts to generate up to $N_{tours}$ number of initial feasible tours within a prescribed initial search time limit, terminating early if no new feasible tour can be found (line \ref{alg:line:endPhase1:TPBS}). To quickly obtain an initial feasible tour, the high-level search ${\tt HighLevelDFS}$ (Alg \ref{alg:highLevelDFS}), is invoked to compute a candidate GTSP tour with associated edges $\mathcal{E}_{tour}$. Each edge $e \in \mathcal{E}_{tour}$ is then evaluated by ${\tt EvalEdge}$ (Alg \ref{alg:EvalEdge}). If all edges are found to be feasible after evaluation, the tour is feasible, and hence added to $\mathcal{F}_{init}$. Additionally, to encourage diversity between the current and the next feasible tour, the time window transitions associated with the current feasible tour are penalized. For each edge $e =(v,v') \in \mathcal{E}_{tour}$, ${\tt WinEdges}$ returns all edges in $\mathcal{E}_{H}$ connecting nodes with the same associated time window as $v$ to nodes with the same associated time window as $v'$, denoted by $\mathcal{E}_{win}$. The penalty associated with each edge in $\mathcal{E}_{win}$ is increased by an amount $\delta$. If one or more edges in $\mathcal{E}_{tour}$ were found to be infeasible after evaluation, ${\tt HighLevelDFS}$ is invoked again to generate a new candidate tour, and the evaluation and subsequent steps are repeated, subject to the Phase 1 termination conditions.

% \vspace{1mm}
{\it Phase 2} of the algorithm is described by lines \ref{alg:line:Phase2start:TPBS}--\ref{alg:line:Phase2end:TPBS}. The initial feasible tours obtained in Phase 1 are first sorted in the increasing order of tour duration $t_f$, and are then improved sequentially over the remainder of the total algorithm time limit. To improve an initial feasible tour $\mathcal{Q}$ it is treated as an incumbent and passed to ${\tt ImproveTour}$ which represents a user-selected GTSP heuristic (e.g., GLNS \cite{smith2017glns}, PGLNS \cite{bhat2025parallel}, or GLKH \cite{helsgaun2015solving}). It then returns a candidate improved GTSP tour $\mathcal{Q}'$ with associated edges $\mathcal{E}_{tour}$. As in Phase 1, all edges in $\mathcal{E}_{tour}$ are evaluated by ${\tt EvalEdge}$. If after evaluation, the cost of $\mathcal{Q}'$ is less than the duration $t_f$ of the incumbent, the incumbent $\mathcal{Q}$ is updated to $\mathcal{Q}'$. Note that the cost of a tour equals its duration $t_f$ if it is feasible and $+\infty$ otherwise. Since the incumbent is always feasible, it can only be updated by another feasible tour of lower cost. Finally, if all the edges in $\mathcal{E}_{tour}$ were already evaluated previously (i.e., line \ref{alg:line:uneval:EvalEdge} in Alg \ref{alg:EvalEdge} is not satisfied each time ${\tt EvalEdge}$ is called in line \ref{alg:line:eval2:TPBS} of Alg \ref{alg:TPBS}), the termination criteria for tour improvement is reached, and $\mathcal{Q}$ is added to $\mathcal{F}_{imp}$. Otherwise, $\mathcal{Q}$ is passed again to ${\tt ImproveTour}$ and the subsequent steps are repeated, subject to the Phase 2 termination conditions.
After both phases are complete, Alg \ref{alg:TPBS} picks the lowest cost tour $\mathcal{Q}^{*}$ from $\mathcal{F}_{init} \cup \mathcal{F}_{imp}$ with associated duration $t^{*}_f$, and finds an associated agent trajectory $\tau^{*}_a$ by calling ${\tt GenAgentTraj}$ (Alg. \ref{alg:GenAgTraj}). Finally, the best feasible solution found, $\mathcal{Q}^{*}, \tau^{*}_a, t^{*}_f$ is returned.

\begin{algorithm}[t!]
    \label{alg:EvalEdge}
    \DontPrintSemicolon
    \SetKw{Pass}{pass}
    \caption{${\tt EvalEdge}$ $(e, \mathcal{G}_{L}, \mathcal{G}_{H})$}
    \If{\textnormal{$e$ is not marked as evaluated}}{\label{alg:line:uneval:EvalEdge}
    $e = (\bar{\pi}_i(t^i), \bar{\pi}_j(t^j))$ \;
    \lIf {${\tt StraightLineFeasible}$ $(\bar{\pi}_i(t^i), \bar{\pi}_j(t^j))$} {$\textnormal{mark $e$ as feasible}$}
    \lElse{$\gamma_e \gets$ ${\tt LowLevelDFS}$ $(\mathcal{G}_{L}, \bar{\pi}_i(t^i), \bar{\pi}_j(t^j))$}
    \If {$\gamma_e \textnormal{ is NULL }$} 
    {\textnormal{Mark $e$ as infeasible} \;
    $\mathcal{G}_{H} \gets$ ${\tt UpdateCost}$ $(+\infty, e, \mathcal{G}_{H})$ \;}
    \lElse{\textnormal{Mark $e$ as feasible}}
    \textnormal{Mark $e$ as evaluated} \;
    }
\end{algorithm}

\begin{algorithm}[t!]
    \label{alg:GenAgTraj}
    \DontPrintSemicolon
    \caption{${\tt GenAgentTraj} (\mathcal{Q},\mathcal{G}_{L})$}
    $\gamma \gets ()$ \;
    $\mathcal{E}_{tour} \gets {\tt TourEdges} (\mathcal{Q})$ \;
    \For{$\textnormal{each edge } e \in \mathcal{E}_{tour}$}{
     $e = (\bar{\pi}_i(t^i), \bar{\pi}_j(t^j))$ \;
    \lIf {${\tt StraightLineFeasible}$ $(\bar{\pi}_i(t^i), \bar{\pi}_j(t^j))$} {$\gamma_e \gets (\bar{\pi}_i(t^i), \bar{\pi}_j(t^j))$}
    \lElse{$\gamma_e \gets$ ${\tt LowLevelDFS}$ $(\mathcal{G}_{L}, \bar{\pi}_i(t^i), \bar{\pi}_j(t^j))$}
    $\gamma \gets {\tt Extend}(\gamma, \gamma_e)$ \;
    }
    \Return{${\tt Traj} (\gamma)$}
\end{algorithm}

% \vspace{1mm}
We will now briefly explain subroutines ${\tt HighLevelDFS}$, ${\tt EvalEdge}$, and ${\tt GenAgentTraj}$. ${\tt HighLevelDFS}$ (Alg \ref{alg:highLevelDFS}) seeks to find a GTSP tour which starts at $\mathcal{C}_s = \{\bar{\pi}_s(0)\}$, visits every cluster $\mathcal{C}_i$ for $i \in \mathcal{T}$ exactly once, before visiting cluster $\mathcal{C}_{s'}$. This achieved using a DFS-style search like in \cite{bhat2025parallel}. The search precomputes a set BEFORE[$v$] for each $v \in \mathcal{V}_{H}$ containing all targets whose clusters cannot be feasibly visited after $v$, and explores \emph{search nodes} denoted by a tuple $u = (\mathcal{U},v)$ with $\mathcal{U} \subseteq \mathcal{T}'$ and $v \in \mathcal{V}_{H}$. A search node represents the set of paths in $\mathcal{G}_{H}$ that visits all clusters corresponding to targets in $\mathcal{U}$ (in any order) before terminating at node $v$. A backpointer is maintained for each search node $u$, denoted as $u$.bp, equal to a search node previously popped from the stack. If $\mathcal{U} = \mathcal{T}'$ (independent of order), a tour $\mathcal{Q}$ is recontructed from $u$. If $\mathcal{Q}$ was not previously added to $\mathcal{F}_{init}$, the search terminates, and $\mathcal{Q}$ is returned. Otherwise, the search continues (lines \ref{alg:line:termstart:highDFS}--\ref{alg:line:termend:highDFS}). If $\mathcal{U} \neq \mathcal{T}'$, a set of \emph{successor $\mathcal{G}$-nodes} $v' \in \mathcal{G}_{H}$ are generated (line~\ref{alg:line:succ:highDFS}) that satisfies (1) Edge $e=(v,v')$ is not marked infeasible, (2) $v' \notin \mathcal{C}_i$ for any $i \in \mathcal{U}$, and (3) BEFORE[$v'$] $\subseteq \mathcal{U}$. Condition (1) ensures a feasible tour in $\mathcal{G}_{H}$, (2) ensures that each cluster is visited once, and (3) ensures that by visiting $v'$, we do not prevent any unvisited clusters from being visited. For each successor $\mathcal{G}$-node $v'$, we generate a successor search node $u'=(\mathcal{U} \cup \{{\tt target}(v')\}, v')$ where ${\tt target}(v')$ denotes the target (or depot) corresponding to $v'$ in $\mathcal{T}'$. Successor search nodes $u'$ are pushed to the stack in order of decreasing ${\tt Cost} (e,\mathcal{G}_{H}) + \mathcal{P}_{H}[e]$, where $e=(v,v')$, such that the least-cost successor gets popped from the stack next.

% \vspace{1mm}
${\tt EvalEdge}$ (Alg. \ref{alg:EvalEdge}) lazily evaluates a previously unevaluated edge $e \in \mathcal{E}_{H}$. It first checks whether a straight-line trajectory connecting the endpoints of $e$ intersects any obstacle over time. If not, $e$ is marked feasible. Otherwise, ${\tt LowLevelDFS}$ is invoked on $\mathcal{G}_{L}$ to search for a collision-free path connecting the endpoints. If no such path exists, $e$ is marked infeasible and its cost in $\mathcal{G}_{H}$ is set to $+\infty$. In all cases, $e$ is marked as evaluated.
${\tt GenAgentTraj}$ (Alg \ref{alg:GenAgTraj}) constructs a continuous agent trajectory from a feasible high-level tour $\mathcal{Q}$. For each edge $e$ in the tour, it attempts to generate a local path $\gamma_e$ connecting the edge endpoints as follows: If a straight-line trajectory connecting the endpoints is collision-free, we set $\gamma_e = e$. Otherwise, $\gamma_e$ is computed by invoking ${\tt LowLevelDFS}$ on $\mathcal{G}_{L}$. The construction of $\mathcal{G}_{L}$ ensures time-feasible, collision-free trajectory segments connecting consecutive points in $\gamma_e$. The local paths for all tour edges are concatenated in tour order to form a complete space–time path $\gamma$, which is finally converted into a continuous agent trajectory via interpolation.

\section{NUMERICAL RESULTS}

\begin{comment}
\begin{figure}[t]
    \centering
    \includegraphics[width=1\linewidth]{figures/SuccessRT.png}
    \caption{Success rates and average runtimes for all approaches. \\
    (a)–(d) Experiment 1: increasing obstacle count from $5$--$40$ with fixed total time window duration of $60 \, secs$. \\
    (e), (c), and (f) Experiment 2: varying total time window duration from $40$--$80$ $secs$ with 20 obstacles.}
    \label{fig:successRT}
\end{figure}
\end{comment}

We ran two experiments on a laptop with an Intel Core I7-7700HQ 2.80GHz CPU, and 16GB RAM.
The implementation was mainly in Python 3.11.6 for all approaches, with $\tt{HighLevelDFS}$ implemented in C++, and $\tt{ImproveTour}$ using PGLNS \cite{bhat2025parallel} since it accepts an initial tour. We set the number of threads to 1 so it works the same as GLNS \cite{smith2017glns}. Additionally, we let PGLNS terminate early if it found an improving tour containing unevaluated edges, before reaching GLNS termination criterion. The MICP was solved using Gurobi 10.0.3 \cite{gurobi}.
The first experiment varied the number of obstacles to be $5, 10, 20$, and $40$ while keeping the total time window duration per target fixed at $60 \, secs$, and the second experiment varied the total target time window duration to be $40, 60$, and $80 \, secs$ while keeping the number of obstacles fixed at $20$. For both experiments, the number of targets were varied to be $5, 10, 20$, and $40$, with each target assigned two time windows of equal duration.

% \vspace{1mm}
We randomly generated instances where targets follow nonlinear trajectories, and obstacle trajectories are kept linear. Target trajectories were obtained by fitting cubic B-splines through piecewise-linear trajectories with speeds drawn randomly from $[0.25, 0.5] \, units/sec$. Obstacle speeds were also drawn randomly from this same range, and the agent's maximum speed was set to $4 \, units/sec$. All targets and obstacles were confined within a square workspace of size $40 \, units$. The obstacles were chosen to be squares with size randomly drawn from $[2,3] \, units$, and the depot location was fixed at the top right corner of the workspace. To avoid infeasible cases, we applied a conservative feasibility screening step, retaining only instances for which a feasible tour could be identified by Phase-1 of TPBS under fully relaxed time window constraints. Based on the corresponding trajectories, time windows were then defined so as to preserve feasibility of the generated instances.

% \vspace{1mm}
We compare our approaches against a baseline method introduced in \cite{li2023dynamic}. The method first attempts to solve the high-level GTSP on $\mathcal{G}_{H}$ using a heuristic. If the straight-line trajectories corresponding to the tour edges are collision-free, the solution is returned. Otherwise, two alternative strategies are invoked. The first strategy repeatedly solves the GTSP while marking colliding tour edges as infeasible, until a feasible solution is found or a time limit is reached. The second strategy is similar but additionally attempts to bypass colliding obstacles by adding space-time points around them at the time of collision, to clusters that must now be visited by the GTSP tour. The lower-cost solution from the two strategies is then returned. We ran the baseline using GLKH \cite{helsgaun2015solving} (like in \cite{li2023dynamic}), as well as GLNS, as the GTSP heuristic.

% \vspace{1mm}
For both experiments, we generated 20 instances for each combination of the number of targets, number of obstacles, and total time-window duration. Each solver was run on all instances with a $150s$ time limit, and we report the success rates, and average runtimes, and cost ratios, where success rate denotes the fraction of instances for which a feasible solution is returned, and the cost ratio for a solver is defined as $(cost_{\text{solver}} - cost_{\text{TPBS}})/cost_{\text{TPBS}}$, averaged over all instances for which both methods return feasible solutions. Additionally, we report the average best solution costs and total runtimes for Phase~1 and Phase~2. For all solvers, we used a common discretization of time at $1 \, sec$ intervals, and for the construction of $\mathcal{G}_{L}$ in TPBS, we uniformly sampled the workspace with $1 \, unit$ spacing in both the $x$ and $y$ directions. We set the penalty parameter $\delta$ for TPBS to $5$, parameter $N_{tours}$ to $4$, and allocated half of the total $150s$ time budget to Phase~1. 
% For ${\tt ImproveTour}$, we used PGLNS \cite{bhat2025parallel} since it accepts an initial tour. We set the number of threads to 1 so it works the same as GLNS.

% \begin{comment}
\begin{figure}[t!]
    \centering
    \includegraphics[width=1\linewidth]{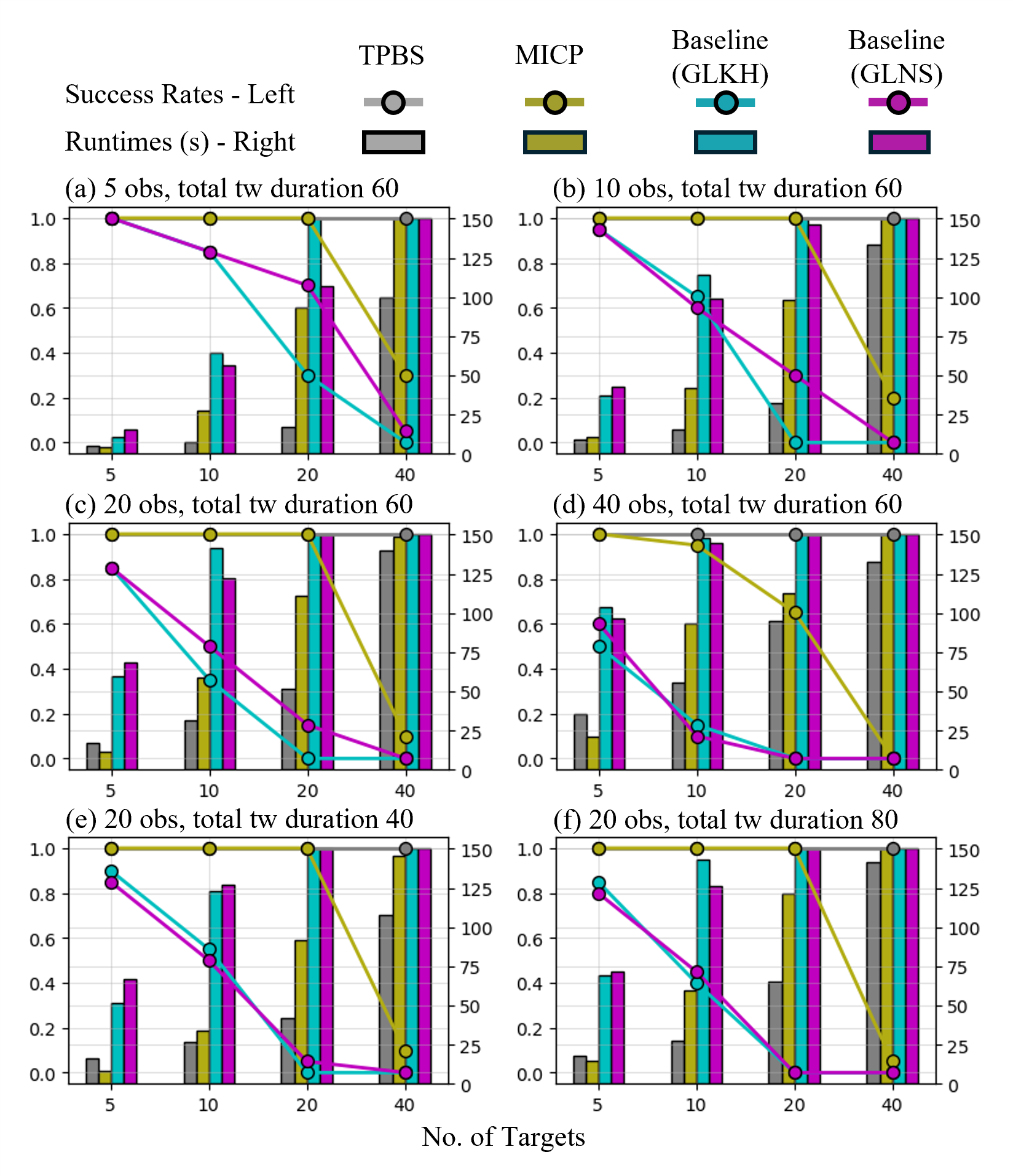}
    \caption{Success rates and average runtimes for all approaches. \\
    (a)–(d) Experiment 1: increasing obstacle count from $5$--$40$ with fixed total time window duration of $60 \, secs$. \\
    (e), (c), and (f) Experiment 2: varying total time window duration from $40$--$80$ $secs$ with 20 obstacles.}
    \label{fig:successRT}
\end{figure}
% \end{comment}

\begin{figure}[t!]
    \centering
    \includegraphics[width=1\linewidth]{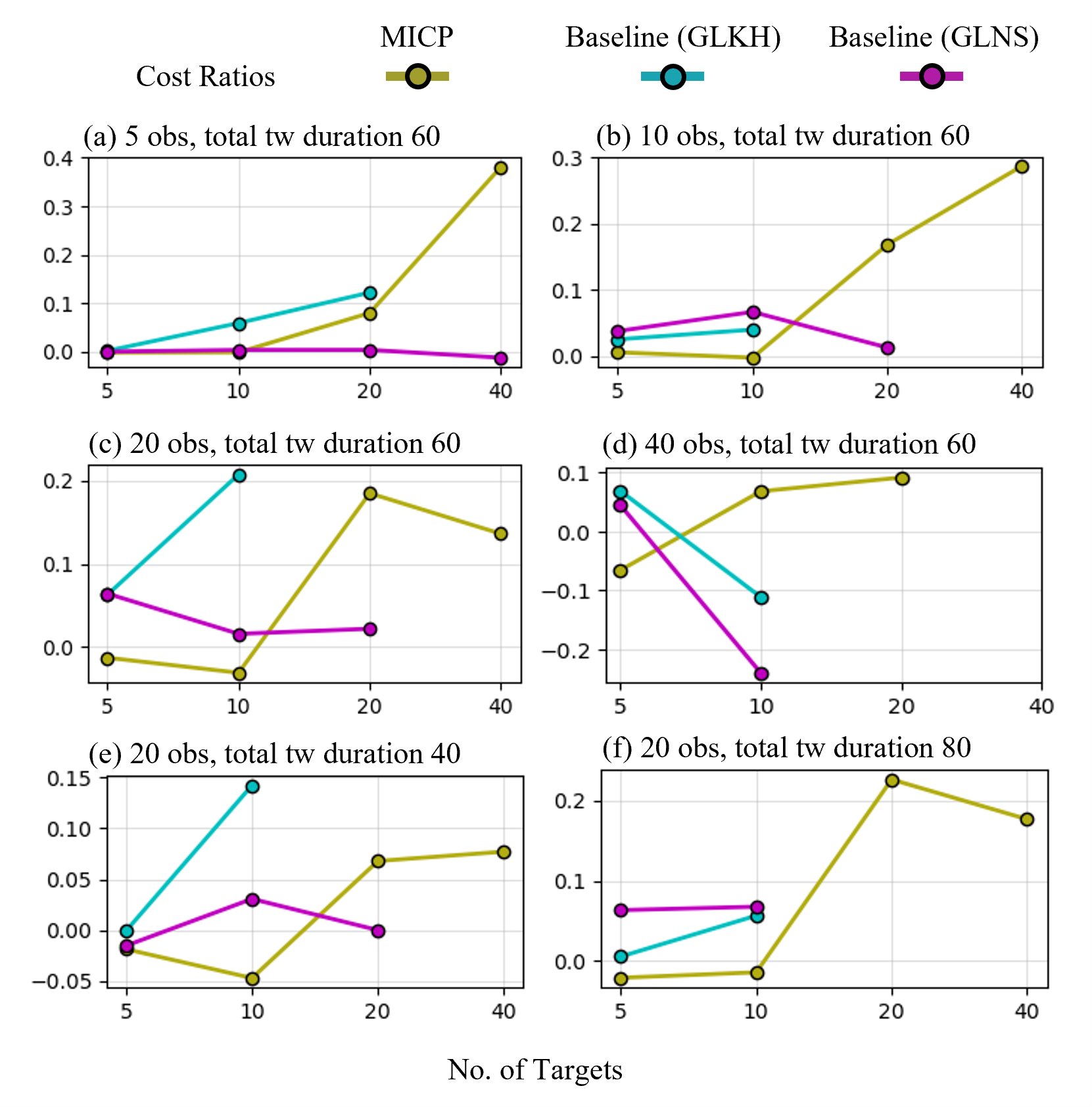}
    \caption{Average cost ratios of MICP and both baselines with respect to TPBS for Experiments 1 and 2 (same layout as Fig.~\ref{fig:successRT}).}
    \label{fig:costRatio}
\end{figure}

\begin{figure}[t!]
    \centering
    \includegraphics[width=1\linewidth]{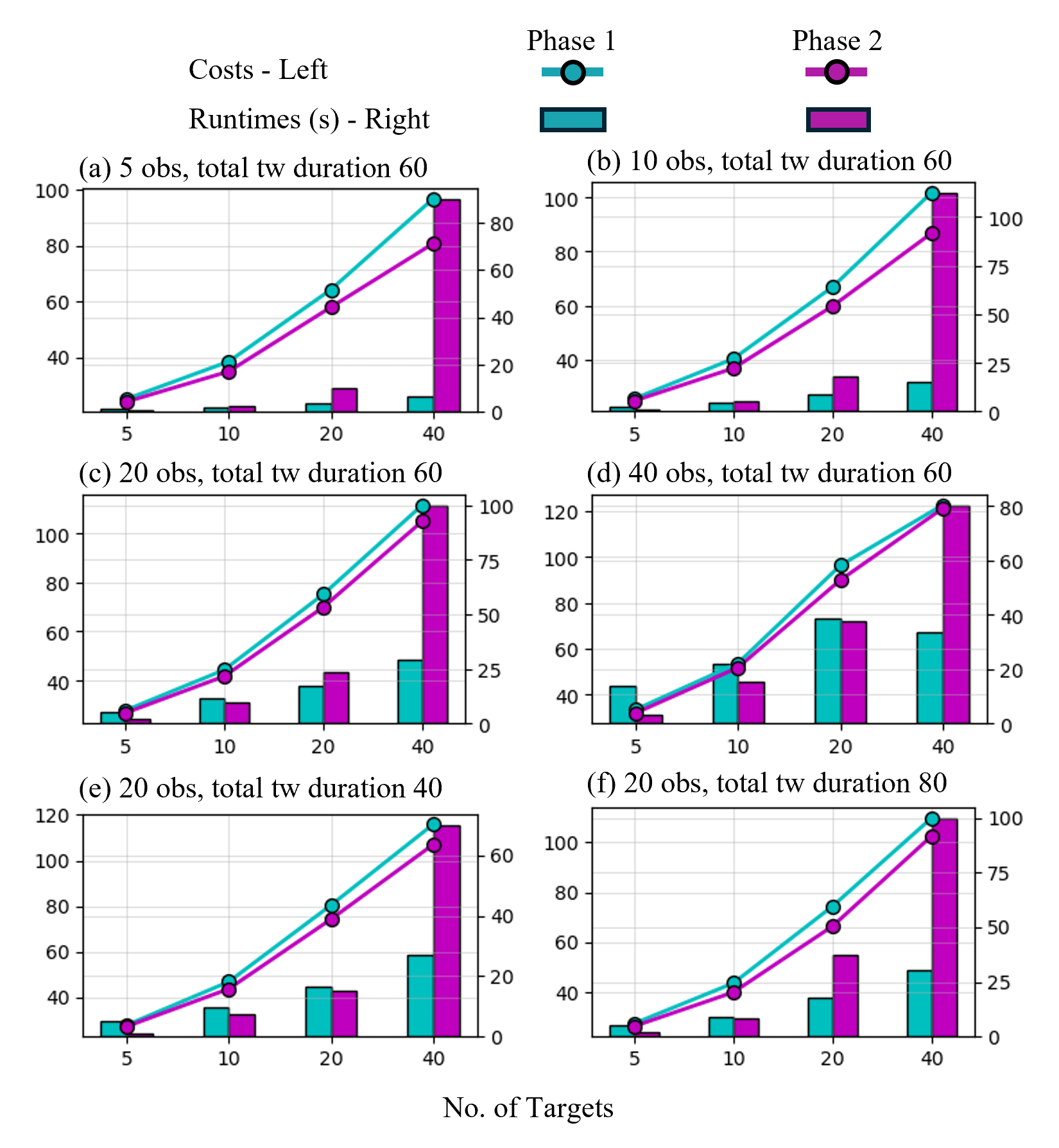}
    \caption{Average of best solution costs and total runtimes for Phase~1 and Phase~2 of TPBS for Experiments 1 and 2 (same layout as Fig.~\ref{fig:successRT}).}
    \label{fig:twoPhase}
\end{figure}

\subsection{Experiment 1: Varying Number of Obstacles}
In this experiment, we vary the number of obstacles to be $5, 10, 20, 40$ while keeping the total time window duration for each target at $60 \, secs$.
The success rates and runtimes are shown in Fig.~\ref{fig:successRT} (a)--(d). The proposed approaches achieve significantly higher success rates than the baselines across all tested instances. In particular, TPBS successfully solved all tested instances, while the MICP formulation remained consistently successful only up to $20$ targets and obstacles. For $40$ targets, the success rate of MICP drops sharply for all obstacle counts, with further degradation observed as the number of targets increases at $40$ obstacles. The success rates of the baseline methods decrease as the number of targets increases, across all obstacle counts. This degradation becomes more pronounced as the number of obstacles grows. We observe that the baseline performs better when using GLNS as the GTSP heuristic compared to GLKH; however, for $20$ targets and obstacles and beyond, both variants exhibit very low success rates, with GLNS achieving only a $20\%$ success rate at $20$ targets and obstacles.
We observe that TPBS consistently achieves the fastest runtime and completes well within the $150s$ limit, particularly for instances up to $20$ targets and obstacles. For $5$ targets, MICP is marginally faster than TPBS in most cases, except at $10$ obstacles. However, as problem size increases, TPBS significantly outperforms MICP, typically requiring half or less of the runtime for up to $20$ targets and obstacles. At $40$ targets, TPBS is the only method that reliably produces feasible solutions within the time limit, while all others frequently exhaust the runtime budget. Among the baselines, GLNS is consistently faster than GLKH, but for $10$ obstacles and above, both typically reach the time limit for $20$ or more targets, reflecting their low success rates. For $5$ and $10$ targets, the baseline methods require roughly twice the runtime of MICP, underscoring their limited scalability.

The cost ratios are shown in Fig.~\ref{fig:costRatio} (a)--(d). For $5$ and $10$ obstacles, all competing methods exhibit mainly nonnegative cost ratios across all target counts, indicating no significant improvement over TPBS. At $20$ obstacles, MICP achieves a modest $\approx5\%$ improvement for $10$ targets. For $40$ obstacles, isolated improvements are observed: MICP yields close to $10\%$ lower cost for $5$ targets, while GLKH and GLNS outperform TPBS for $10$ targets by approximately $10\%$ and $25\%$, respectively. In all remaining cases, TPBS achieves comparable or lower cost.

Finally, Fig.~\ref{fig:twoPhase} (a)--(d) reports the best solution costs and total runtimes for Phase~1 and Phase~2. We observe that cost improvements from Phase~2 generally increase with the number of targets, but diminish as the number of obstacles grows. In particular, for $40$ obstacles, further improving Phase~1 solutions becomes difficult, as even finding feasible solutions is challenging. The runtimes of both phases increase with problem size, scaling with both the number of targets and obstacles. For instances with $40$ targets, Phase~1 is substantially faster than Phase~2, requiring less than half the runtime across all obstacle counts.

\subsection{Experiment 2: Varying Total Time Window Duration}
In this experiment, we vary the total time window duration per target to be $40, 60, 80 \, secs$, while fixing the number of obstacles constant at $20$. Like in the previous experiment, success rates and runtimes are reported in Fig.~\ref{fig:successRT}, cost ratios in Fig.~\ref{fig:costRatio}, and the best solution costs and total runtimes for Phase~1 and Phase~2 in Fig.~\ref{fig:twoPhase}. Across all figures, we reuse subfigure (c) for the $60 \, secs$ case. Additionally, subfigures (e) and (f) corresponds to $40$ and $80 \, secs$ respectively. From Fig.~\ref{fig:successRT}, we observe that the runtimes of all approaches increase with longer time window durations, particularly for larger numbers of targets. Moreover, the success rates of the proposed methods remain largely stable, whereas those of the baseline approaches fluctuate as the time window duration increases. From Fig.~\ref{fig:costRatio}, we observe broadly consistent trends in cost ratios across all methods, with an increase as the time-window duration increases for the MICP. Interestingly, the baseline with GLKH attains a smaller cost ratio at $80 \, secs$, than it does for shorter durations. Finally, from Fig.~\ref{fig:twoPhase}, we observe a slight increase in runtimes and consistent cost improvements from Phase~2, with longer time window durations.

\section{CONCLUSION AND FUTURE WORK}
In this paper, we presented a Mixed-Integer Conic Programming formulation and a Two-Phase Bilevel Search algorithm for the Moving-Target Traveling Salesman Problem with Moving Obstacles. Extensive numerical experiments demonstrate that the proposed approaches significantly outperform a baseline method across a broad range of problem instances. Promising directions for future work include incorporating dynamic constraints on the agent, extending to multi-agent scenarios, and developing fast algorithms with provable guarantees on completeness and optimality.

\addtolength{\textheight}{-12cm}   % This command serves to balance the column lengths

%%%%%%%%%%%%%%%%%%%%%%%%%%%%%%%%%%%%%%%%%%%%%%%%%%%%%%%%%%%%%%%%%%%%%%%%%%%%%%%%

%%%%%%%%%%%%%%%%%%%%%%%%%%%%%%%%%%%%%%%%%%%%%%%%%%%%%%%%%%%%%%%%%%%%%%%%%%%%%%%%

%%%%%%%%%%%%%%%%%%%%%%%%%%%%%%%%%%%%%%%%%%%%%%%%%%%%%%%%%%%%%%%%%%%%%%%%%%%%%%%%

\begin{comment}
\section*{APPENDIX}

Appendixes should appear before the acknowledgment.

\section*{ACKNOWLEDGMENT}
\end{comment}

\begin{comment}
The preferred spelling of the word ÒacknowledgmentÓ in America is without an ÒeÓ after the ÒgÓ. Avoid the stilted expression, ÒOne of us (R. B. G.) thanks . . .Ó  Instead, try ÒR. B. G. thanksÓ. Put sponsor acknowledgments in the unnumbered footnote on the first page.
\end{comment}

%%%%%%%%%%%%%%%%%%%%%%%%%%%%%%%%%%%%%%%%%%%%%%%%%%%%%%%%%%%%%%%%%%%%%%%%%%%%%%%%

\begin{comment}
References are important to the reader; therefore, each citation must be complete and correct. If at all possible, references should be commonly available publications.
\end{comment}

\bibliographystyle{IEEETran}
\bibliography{REFERENCES}

\end{document}